\documentclass[12pt,journal,compsoc,onecolumn]{IEEEtran}
\usepackage{amsmath,graphicx,cleveref,caption,subcaption,dsfont,multirow,multicol,amssymb,amsfonts,mathtools,pifont}
\usepackage{amsmath,epsfig}
 \usepackage{lingmacros}
\usepackage{tree-dvips}
\usepackage{amsmath} 
\usepackage{amssymb}
\usepackage{pifont}
\usepackage[noadjust]{cite}

\newcommand{\cmark}{\ding{51}}
\newcommand{\xmark}{\ding{55}}

\def\x{{\mathbf x}}

\def\XX{{\bf X}}

\def\Y{{\cal Y}}
\def\S{{\cal I}}

\def\x{{\bf x}}
\def\y{{\bf y}}
\def\D{{\cal D}}

\def\F{{ f}}
\def\DD{{d}}
\def\tr{{\bf tr}}
\def\1{{\bf 1}}
\def\I{{\cal I}} 
\def\p{{p}}
\def\q{{q}}
\def\x{{\bf x}}  
\def\Y{{\cal  Y}}  
\def\V{{\bf V}}
\newtheorem{proposition}{Proposition}

\usepackage[ruled,vlined]{algorithm2e}

\title{Frugal Learning of Virtual Exemplars  for Label-Efficient Satellite Image Change Detection}

\author{ Hichem Sahbi$^1$ \ \ \ \ \ \  \ \ \ \ \ \  \ \ \ \ \ \   \ \ \ \ \ \ Sebastien Deschamps$^{1,2}$ \\ $ $ \\ 
$^1$Sorbonne University, UPMC, CNRS, LIP6, France \\ $^2$Theresis Thales, France}

\begin{document}
\maketitle

\begin{abstract}
In this paper, we devise a novel interactive  satellite image change detection algorithm based on active learning. The proposed framework is  iterative and relies on a question \& answer model which asks the oracle (user) questions about  the most informative display (subset of critical images), and according to the user's responses, updates  change detections.  The contribution of our framework resides in a novel display model which selects  the most representative and diverse  virtual exemplars that adversely challenge the  learned  change detection functions, thereby leading to highly discriminating functions in the subsequent  iterations of active learning.   Extensive experiments, conducted on the challenging task of interactive satellite image change detection, show the superiority of the proposed  virtual display model against the related work.  \\

\end{abstract}

\section{Introduction}
\label{sec:intro}
Satellite image change detection aims at localizing instances of {\it relevant}  changes in a given scene acquired at different instants.  This problem has many applications including  evaluation of damaged infrastructures in order to prioritize  rescue and  disaster response after natural hazards (tornadoes, earthquakes, etc) \cite{ref4,ref5}.  This task is also challenging as relevant changes are diverse and scenes are  subject to many irrelevant changes due to sensor artefacts,  registration errors, illumination variations,  occlusions,  weather conditions,  etc.   Early change detection solutions  were  based on simple comparisons of multi-temporal signals, via image differences and thresholding,  using vegetation indices, principal component and change vector analyses \cite{ref7,ref9,ref11,ref13}.  Other methods either require a preliminary preprocessing step that attenuates  the effects of irrelevant changes (by correcting radiometric variations, occlusions,  shadows, and by finding the parameters of sensors for registration \cite{ref14,ref15,ref17,ref20}) or consider these effects   as  a part of appearance modeling  \cite{ref21,ref25,ref26,ref27,ref28}. \\

\indent Among the aforementioned  methods, those based on machine learning are particularly successful, but their accuracy is highly dependent on the availability of large collections of  labeled training data.  Indeed,   these approaches are limited by the scarcity of labeled data in order to comprehensively capture the huge  variability in relevant and irrelevant changes.  Besides,  even when larger training collections are available, their labeling may not reflect the user's subjectivity and intention.  Many existing solutions try to overcome these limitations by making training frugal and less dependent on large collections of labeled data. These solutions include few shot \cite{reff45} and self-supervised  learning \cite{refff2}; however,   these approaches are oblivious to the users' intention.  Alternative solutions,  based on active learning  \cite{reff1,reff2,reff16,reff15,reff53,reff12,reff74,reff58,reff13},  are rather preferred where users annotate very few examples of relevant and irrelevant changes according to their intention,  prior to retrain  user-aware   change detection functions. \\

\indent In this paper, we introduce a novel  interactive satellite image change detection algorithm based on a question \& answer model that queries the  intention of the user (oracle), and updates change detections  accordingly.    The oracle frugally provides these annotations   only on the most informative displays which are {\it learned} instead of being directly sampled from the {\it fixed} pool of unlabeled data.   A conditional probability distribution is defined which measures  the relevance of  each exemplar in the learned displays given the pool of unlabeled data. This conditional distribution is  learned  using a novel adversarial criterion that finds the most diverse,  representative,  and uncertain exemplars which challenge (the most) the previously trained change detection functions.   Note that,  in spite of being adversarial,   the framework proposed in our display model  is conceptually different from generative adversarial networks (GANs) \cite{refffabc}. Indeed,  GANs seek to generate fake data that mislead the trained discriminators while our proposed model aims at generating the most critical data for further annotations; i.e., the most representative and diverse  exemplars which increase the uncertainty,  and challenge (the most) the current discriminator,  and ultimately lead to more accurate classifiers  in the subsequent iterations of  change detection.  Experiments, conducted on the challenging task of interactive satellite image change detection, show the relevance of our {\it exemplar and display learning} model against the related work.  

\section{Proposed method}
\label{sec:format}
Let $\I_r = \{\p_1, \dots , \p_n\}$, $\I_t = \{\q_1, \dots , \q_n\}$  denote two registered satellite images captured at two instants $t_0$, $t_1$, with $\p_i$, $\q_i \in \mathbb{R}^d$. Considering  $\I = \{\x_1,\dots, \x_n\}$, with each $\x_i$ being an aligned patch pair $(\p_i, \q_i)$, and $\Y = \{\y_1, \dots, \y_n\}$ the underlying unknown labels; our goal is to train a change detection function  $f: {\cal I} \rightarrow \{-1,+1\}$ that finds the unknown labels in $\{\y_i\}_i$ with $\y_i = +1$ if $\q_i \in  \I_t$  corresponds to a ``change'' w.r.t. $\p_i \in \I_r$, and $\y_i = -1$ otherwise. Designing $f$ requires a subset of training data annotated by an oracle. As these annotations are highly expensive, building $f$ should be accomplished with as few annotations as possible while maximizing accuracy.   
\subsection{Interactive satellite image change detection}
Our change detection algorithm is interactive; it is based on a question \& answer model which shows an informative subset of images (referred to as display) to an oracle, gathers their annotations and trains a decision criterion $f$ accordingly. Considering $\D_t \subset \I$ as a display shown to the oracle at iteration $t$, and $\Y_t$ as the unknown labels of $\D_t$; in practice $|\D_t|$ and the maximum number of iterations (denoted as $T$) are set depending on a fixed annotation budget. \\  Starting from a random display $\D_0$, we build our change decision criteria iteratively while running the following steps for $t=0,\dots,T-1$ \\
\noindent 1/ Query the oracle about the labels of $\D_t$ and train a decision function $f_t (.)$ on $\cup_{k=0}^t (\D_k,\Y_k)$; in our experiments,  $\{f_t (.)\}_t$ correspond to max-margin classifiers built on top of convolutional features.\\
2/ Select the next display $\D_{t+1}\subset \S\backslash\cup_{k=0}^t \D_k$ to show to the oracle; it is clear that a brute force strategy that considers all the possible displays $\D \subset \S\backslash\cup_{k=0}^t \D_k$, learns the underlying classifiers $f_{t+1} (.)$ on $\D \cup_{k=0}^t \D_t$ and evaluates their accuracy is highly combinatorial. Display selection strategies, related to active learning, are instead preferred and make display learning more tractable. However,  the design of display selection strategies should be carefully achieved as many of these heuristics are equivalent to (or worse than) basic display strategies that choose data uniformly randomly (see for instance \cite{refff1} and references therein). \\ Our proposed display model in this paper is different from usual sampling strategies (see e.g.  \cite{refff33333}) and relies on synthesizing exemplars (also referred to as virtual exemplars) that maximize diversity, representativity as well as uncertainty. Diversity aims at designing exemplars that allow exploring different modes of $f_{t+1} (.)$ whereas representativity makes those exemplars resembling as much as  possible to the input data. Finally, ambiguity seeks to locally refine the boundaries of the learned decision function $f_{t+1} (.)$.  All the details of our proposed display model are shown in the subsequent section which constitutes the main contribution of this work.
\subsection{Virtual display model}
\indent We consider  a  framework that assigns for each sample $\x_i$ a  conditional distribution $\{\mu_{ik}\}_{k=1}^K$ measuring the probability of assigning $\x_i$ to each of the K-virtual exemplars, and the latter constitute the subsequent display  $\D_{t+1}$. In contrast to our previous work \cite{refff33333}, the proposed method in this paper neither   requires  hard thresholding nor ranking of the memberships $\{\mu_{ik}\}_{k=1}^K$ in order to define  $\D_{t+1}$; instead, entries of $\D_{t+1}$, also denoted as $\V$,  together with $\{\mu_{ik}\}_{k=1}^K$,  are found by minimizing the following constrained objective function  
{
\hspace{-3.5cm}\begin{equation}\label{eq0}
\begin{array}{ll}
\displaystyle   \min_{\V; \mu \geq 0; \mu \1_K = \1_n}    & \displaystyle   \tr\big (\mu \ \DD(\V,\XX)' \big)  \  + \ \alpha  \ [\1'_n  \mu] \log [\1'_n \mu]' \\
                 &  \  +  \  \beta \ \tr\big(\F(\V)' \  \log \F(\V)\big) \ + \ \gamma \ \tr(\mu' \log \mu), 
\end{array}
\end{equation}}
\noindent  here $'$ is the matrix transpose operator,  $\1_{K}$, $\1_{n}$ denote two vectors of $K$ and $n$ ones respectively,  $\mu \in \mathbb{R}^{n \times K}$ is a learned matrix which provides the membership of each input sample $\x_i$  to the k-th virtual exemplar and $\log$ is applied entrywise. In the above objective function, $\DD(\V,\XX) \in \mathbb{R}^{K \times n}$ is the euclidean distance matrix between the virtual exemplars in $\V$ and the original data in $\XX$ while $\F(\V) \in \mathbb{R}^{2 \times K} $ is a scoring matrix whose columns provide the response of the learned decision function $f_t(.) \in [0,1]$ and its complement $1-f_t(.)$ on the K-th virtual exemplars. The first term of the above objective function (rewritten as $\sum_i \sum_k \mu_{ik}  \| \x_i - \V_{k}\|_2^2$) measures the {\it representativity} of the synthesized exemplars $\{\V_{k}\}_k$; it models how close is each training sample $\x_i$ w.r.t its closest (or most representative) $\V_k$, and vanishes when all training samples coincide with their virtual exemplars. The second term (equivalent to  $\sum_{k}   [\frac{1}{n} \sum_{i=1}^n  \mu_{ik}]  \log  [\frac{1}{n} \sum_{i=1}^n  \mu_{ik}]$) captures the {\it diversity} of the generated virtual data as the entropy of the probability distribution of the underlying memberships; this measure is minimal when  training samples are assigned to different virtual exemplars, and vice-versa. The third criterion (rewritten as $\sum_k \sum_c [\F_c(\V)]'_k [\log \F_c(\V)]_k$) measures the {\it ambiguity} (or uncertainty) in $\D_{t+1}$ as the entropy of the scoring function; it reaches its smallest value when virtual exemplars in  $\D_{t+1}$ are evenly scored w.r.t different classes. The fourth term considers that, without any a priori on the three other terms, the membership distribution of each input data is uniform, so its acts as a regularizer and also helps obtaining a closed form solution (as shown subsequently). All these terms are mixed using the coefficients $\alpha, \beta, \gamma \geq 0$. Finally, we consider equality and inequality constraints which guarantee that the membership of each input sample $\x_i$ to the virtual exemplars forms a probability distribution. 
\subsection{Optimization} 
\begin{proposition}
The optimality conditions of Eq.~(\ref{eq0}) lead to the solution
 {
\begin{equation}\label{eq1}
\begin{array}{lll}
  \mu^{(\tau+1)}& :=&\displaystyle  \textrm{\bf diag} \big(\hat{\mu}^{(\tau+1)} \1_K\big)^{-1} \  \hat{\mu}^{(\tau+1)} \\
  \V^{(\tau+1)} &:= & \hat{\V}^{(\tau+1)} \  \textrm{\bf diag} \big(\1'_n {\mu}^{(\tau)} \big)^{-1},
\end{array}  
\end{equation}}
with $\hat{\mu}^{(\tau+1)}$, $\hat{\V}^{(\tau+1)}$ being respectively 
\begin{equation}\label{eq2}
\begin{array}{l}
 \exp\big(-\frac{1}{\gamma}[\DD(\XX,\V^{(\tau)}) + \alpha (\1_n \1'_K + \1_n \log \1'_n \mu^{(\tau)} )]\big)\\
\\ 
\XX \ \mu^{(\tau)}+ \beta \sum_{c} \nabla_v f_c(\V^{(\tau)}) \circ  (\1_d \ [\log f_c(\V^{(\tau)})]' + \1_d \1_K'),
\end{array}
\end{equation}
here $\circ$ is the Hadamard matrix product and $\textrm{\bf diag}(.)$ maps a vector  to a diagonal matrix.                        
\end{proposition} 
Due to space limitation, details of the proof are omitted and  result from the optimality conditions of Eq.~\ref{eq0}'s gradient.  Considering the above proposition, $\hat{\mu}^{(0)}$ and  $\hat{\V}^{(0)}$ are initially set to random values and, in practice, the solution converges to the fixed points (denoted as $\tilde{\mu}$, $\tilde{\V}$) in few iterations. These fixed points define the most {\it relevant} virtual exemplars of $\D_{t+1}$ (according to criterion~\ref{eq0}) which are used to train the subsequent classifier $f_{t+1}$ (see also algorithm~\ref{alg1}).
\begin{algorithm}[!ht]
 
  \KwIn{Images in $\S$, display ${\cal D}_0 \subset {\S}$, budget $T$.}
\KwOut{$\cup_{t=0}^{T-1} (\D_t,\Y_t)$ and $\{f_t\}_{t}$.}
\BlankLine
\For{$t:=0$ {\bf to} $T-1$}{$\Y_t \leftarrow oracle(\D_t)$; \\ 
  $f_{t} \leftarrow \arg\min_{f} {Loss}(f,\cup_{k=0}^t (\D_k,\Y_k))$; \\
 $\tau \leftarrow 0$; $\hat{\mu}^{(0)} \leftarrow \textrm{random}$;  $\hat{\V}^{(0)} \leftarrow \textrm{random}$;\\
Set ${\mu}^{(0)}$ and  ${\V}^{(0)}$ using Eqs.~(\ref{eq1}) and (\ref{eq2});
\BlankLine
 \While{($\footnotesize \|\mu^{(\tau+1)}-\mu^{(\tau)}\|_1 +  \|\V^{(\tau+1)}-\V^{(\tau)}\|_1 \geq\epsilon  \newline \wedge  \tau<\textrm{maxiter})$}{
   Set ${\mu}^{(\tau+1)}$ and  ${\V}^{(\tau+1)}$ using Eqs.~(\ref{eq1}) and (\ref{eq2}); \\  
   $\tau \leftarrow \tau +1$;
 }
$\tilde{\mu} \leftarrow \mu^{(\tau)}$;  $\tilde{\V} \leftarrow \V^{(\tau)}$; \\ 
 ${\D_{t+1} \leftarrow \big\{\x_i \in \S\backslash \cup_{k=0}^t \D_k: \x_i\leftarrow\arg\min_\x\|\x-\V_k\|_2^2 \big\}_{k=1}^K}$.
}
\caption{Display selection mechanism}\label{alg1}
\end{algorithm}
\section{Experiments}
Change detection experiments are conducted on the Jefferson dataset including $2,200$ non overlapping patch pairs (of $30\times 30$ RGB pixels each). These pairs correspond to  registered (bi-temporal) GeoEye-1 satellite images of $2, 400 \times  1, 652$ pixels with a spatial resolution of 1.65m/pixel, taken from the area of Jefferson (Alabama) in 2010 and in 2011. These images show multiple changes due to tornadoes in Jefferson (building destruction, etc.) as well as no-changes (including irrelevant ones as clouds). The ground-truth  consists of 2,161 negative  pairs (no/irrelevant changes) and only 39 positive  pairs (relevant changes), so $>98\%$ of this area correspond to no-changes and this makes the task of finding relevant changes even more challenging. In our experiments, half of the dataset is used to train the display and the learning models while the remaining half for evaluation.  As the two classes (changes/no-changes) are highly imbalanced, we measure  accuracy using the equal error rate (EER) on the eval set. Smaller EER implies better performances.
\subsection{Ablation}
In order to study the impact of different terms of our objective function, we consider them individually, pairwise  and all jointly taken. In this study, the last term of Eq.~\ref{eq0} is always kept as it acts as a regularizer and allows obtaining the closed form in Eq.~\ref{eq1}. The impact of each of these terms and their combination is shown in Table~\ref{tab1}. From these results, we observe the highest impact of representativity+diversity especially at the earliest iterations of change detection, whilst the impact of ambiguity term raises later in order to locally refine the decision functions  (i.e., once the modes of data distribution become well explored). These EER performances are shown  for different sampling percentages defined --- at each iteration $t$ --- as $(\sum_{k=0}^{t-1} |\D_k|/(|\I|/2))\times 100$  with  $|\I|=2,200$ and $|\D_k|$ set to $16$. 
 \begin{table}
 \resizebox{1.0\columnwidth}{!}{
 \begin{tabular}{ccc||cccccccccc||c}
 rep  & div & amb & 1 &2 & 3& 4& 5& 6& 7& 8& 9 & 10 & AUC. \\
 \hline
 \hline
        \xmark  &  \xmark &     \cmark    & 47.81    & 27.29  &  11.15  &  7.97 &   8.18  &  7.31  &  7.97  &  7.94 &   7.50  &  7.90 &  14.10\\
         \xmark  & \cmark  &       \xmark  & 47.81    & 18.72  &  11.24&     7.97 &   8.18&     7.29&     7.59   &  7.88   &  7.50  &   7.90 & 13.21  \\
         \cmark   &  \xmark  &      \xmark &  47.81   &  35.98 &   16.86 &   6.52 &    4.98&     2.67 &    2.03   &  1.80 &    1.45&     1.30 & 12.14 \\
\hline
         \cmark   &      \xmark &   \cmark  &  47.81   &  40.40&   23.86&     9.56&    7.65&     5.75 &    5.47  &   6.12 &    4.40   &  5.72 &  15.67 \\
         \xmark  &  \cmark &   \cmark  &  47.81   & 27.29  &   11.15    & 7.97  &   8.18   &  7.31   &  7.97   &  7.94  &   7.50  &   7.90 & 14.10  \\
         \cmark  & \cmark    &     \xmark  & 47.81   &  29.84 &    17.63 &    6.21  &   4.40   & 2.70  &   1.98 &    1.92  &   1.65   &  1.52 & 11.57 \\
\hline 
        \cmark   &  \cmark  &  \cmark  &  47.81  &  27.61  &   11.76 &    5.74  &   2.95  &   2.39  &   1.89  &   1.61  &   1.55 &    1.34 &\bf10.47 \\
\hline \hline 
\multicolumn{3}{c||}{Samp\%}  & 1.45 &2.90 & 4.36& 5.81& 7.27& 8.72& 10.18& 11.63& 13.09 & 14.54 & - 
\end{tabular}}
 \caption{This table shows an ablation study of our display model. Here rep, amb and div stand for representativity, ambiguity and diversity respectively. These results are shown for different iterations $t=0,\dots,T-1$ (Iter) and the underlying sampling rates (Samp) again defined as $(\sum_{k=0}^{t-1} |\D_k|/(|\I|/2))\times 100$. The AUC (Area Under Curve) corresponds to the average of EERs across iterations.}\label{tab1} 
 \end{table} 
 \subsection{Comparison}\label{compare}
\begin{figure}[tbp]
\centering
\includegraphics[width=0.79\linewidth]{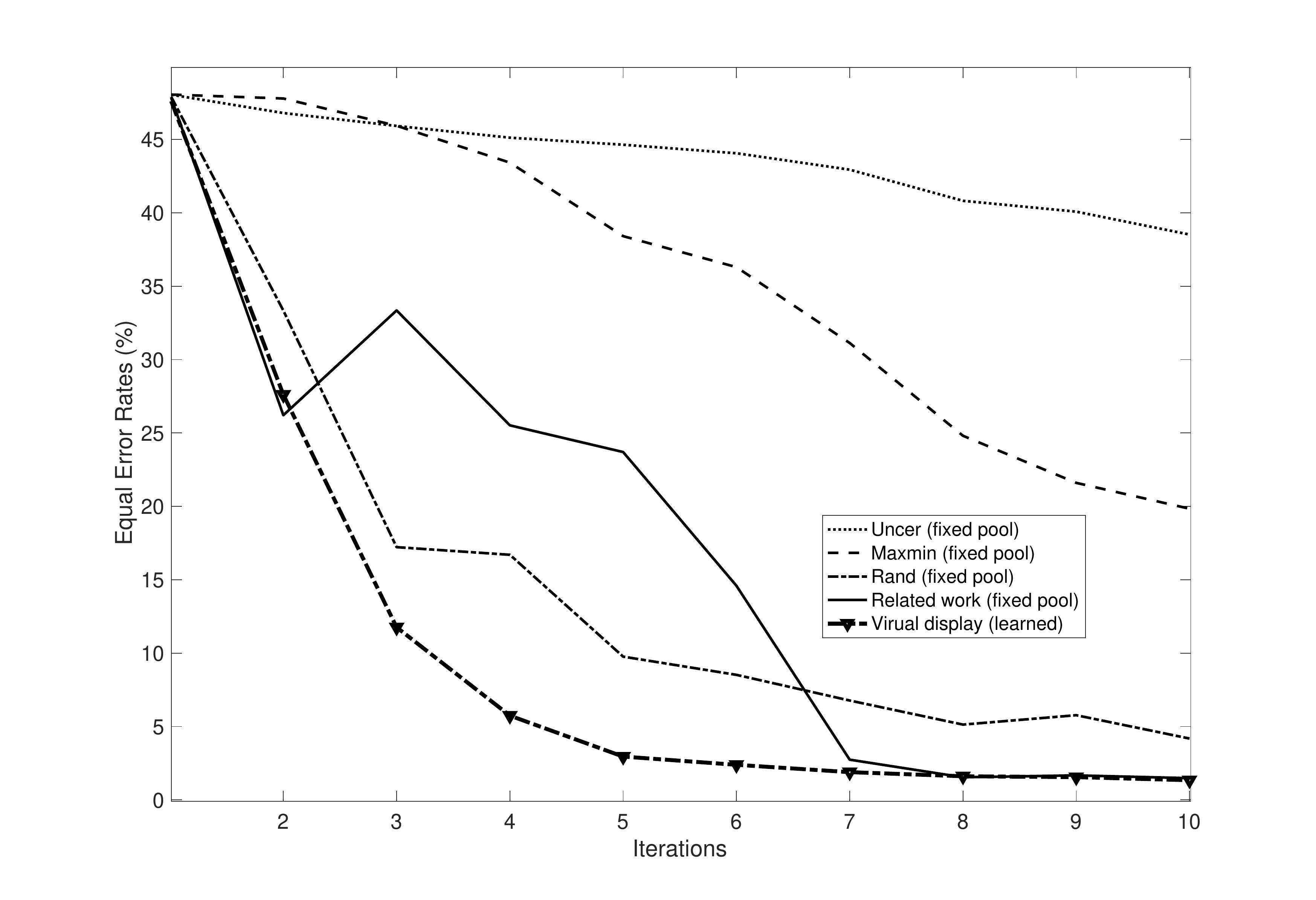}
 \caption{This figure shows a comparison of different sampling strategies w.r.t. different iterations (Iter) and the underlying sampling rates in table~\ref{tab1} (Samp). Here Uncer and Rand stand for uncertainty and random sampling respectively. Note that fully-supervised learning achieves an EER of $0.94 \%$.  Related work stands for the method in \cite{refff33333}; see again section~\ref{compare} for more details.}\label{tab2}\end{figure}

 We further investigate the strength of our display model against other display sampling strategies including {\it random search, maxmin and uncertainty}.  Random selects samples from the pool of unlabeled training data while uncertainty consists in picking, from the same pool, the display whose classifier scores are the most ambiguous (i.e., closest to zero). Maxmin consists in greedily sampling data in $\D_{t+1}$; each sample in  $\x_i \in \D_{t+1} \subset \S\backslash \cup_{k=0}^t \D_k$ is chosen by {\it maximizing its minimum distance w.r.t.  $\cup_{k=0}^t \D_k$},  leading to the most distinct samples in $\D_{t+1}$. We further compare our display model against \cite{refff33333} which consists in assigning a probability measure to the whole unlabeled  set and selecting the display with the highest probabilities.  We also  report performances using the fully supervised setting, as an upper bound, which consists in building a unique classifier on top of the full training set whose annotation is taken from the ground-truth.  EER performances reported in Figure~\ref{tab2} (w.r.t different iterations and sampling rates) show the positive impact of the proposed virtual display model against the aformentioned sampling stratgies. Excepting the model in \cite{refff33333}, most of these comparative methods are powerless to find the (rare) change class sufficiently well. Indeed, both random and maxmin capture the diversity at the early stage of interactive search without being able to refine the decision function at the latest iterations.  In contrast,  uncertainly refines well the decision function but lacks diversity.  The display strategy in \cite{refff33333} gathers the advantages of random and maxmin as well as uncertainty,  but suffers from the rigidity in the selected display (especially at the early iterations) which is taken from a fixed set of training data while our virtual display model is learned and thereby more flexible and   effective at highly frugal regimes.

\section{Conclusion}
We introduce in this paper  a novel interactive change detection method based on active learning.  The strength of the proposed method resides in the flexibility of the learned display model which allows training virtual exemplars. The latter are found while maximizing their diversity and  representativity as well as their ambiguity, leading to an adversarial setting which challenges the current classifier and enhances the subsequent one. 
Experiments, conducted on the task of satellite image change detection, show the outperformance of the proposed virtual display model against different   related models and sampling strategies. 

{

}

\end{document}